\documentclass[lettersize,journal]{IEEEtran}
\usepackage{amsmath,amsfonts,amssymb}
\usepackage{algorithmic}
\usepackage{algorithm}
\usepackage{array}
\usepackage[caption=false,font=normalsize,labelfont=sf,textfont=sf]{subfig}
\usepackage{textcomp}
\usepackage{stfloats}
\usepackage{url}
\usepackage{verbatim}
\usepackage{graphicx}
\usepackage{multirow}
\usepackage{multicol}
\usepackage[square,numbers]{natbib}
\usepackage[table,xcdraw]{xcolor}
\usepackage{colortbl}
\usepackage{utfsym}
\usepackage{epstopdf}

\definecolor{lightred}{rgb}{1, 0.8, 0.8}
\definecolor{lightblue}{rgb}{0.8, 0.8, 1}
\definecolor{lightgreen}{rgb}{0.8, 1, 0.8}
\begin{document}
\title{FA\textsuperscript{3}-CLIP: Frequency-Aware Cues Fusion and Attack-Agnostic Prompt Learning for Unified Face Attack Detection}
\author{
    Yongze Li$^{*}$, 
    Ning Li$^{*}$,
    Ajian Liu$^{\dagger}$,
    Hui Ma, 
    Liying Yang, 
    Xihong Chen, 
    Zhiyao Liang, \\
    Yanyan Liang,~\IEEEmembership{Senior Member,~IEEE},
    Jun Wan,~\IEEEmembership{Senior Member,~IEEE},
    Zhen Lei,~\IEEEmembership{Fellow,~IEEE}
    \thanks{
    Yongze Li, Ning Li, Hui Ma, Liying Yang, Zhiyao Liang and Yanyan Liang are with the School of Computer Science and Engineering, Faculty of Innovation Engineering, Macau University of Science and Technology, Macau, China (e-mail: \{2009853gii30008,2009853dia30001\}@student.must.edu.mo, \{zyliang,yyliang\}@must.edu.mo). 
    
    Ajian Liu, Jun Wan, and Zhen Lei are with the State Key Laboratory of Multimodal Artificial Intelligence Systems (MAIS), Institute of Automation, Chinese Academy of Sciences (CASIA), Beijing 100190, China, and also with the School of Computer Science and Engineering, Faculty of Innovation Engineering, Macau University of Science and Technology, Macau 999078, China (e-mail: \{ajian.liu,jun.wan,zhen.lei\}@ia.ac.cn).

    Xihong Chen is with the School of Software Engineering, Beijing Jiaotong University (BJTU), Beijing 100044, China (e-mail: 22301061@bjtu.edu.cn).

    Corresponding author: Ajian Liu.
}
}
\markboth{Journal of \LaTeX\ Class Files,~Vol.~14, No.~8, August~2021}%
{Shell \MakeLowercase{\textit{et al.}}: A Sample Article Using IEEEtran.cls for IEEE Journals}
\IEEEpubid{0000--0000/00\$00.00~\copyright~2021 IEEE}

\maketitle
\begin{abstract}
	Facial recognition systems are vulnerable to physical (e.g., printed photos) and digital (e.g., DeepFake) face attacks. Existing methods struggle to simultaneously detect physical and digital attacks due to: 1) significant intra-class variations between these attack types, and 2) the inadequacy of spatial information alone to comprehensively capture live and fake cues. To address these issues, we propose a unified attack detection model termed Frequency-Aware and Attack-Agnostic CLIP (FA\textsuperscript{3}-CLIP), which introduces attack-agnostic prompt learning to express generic live and fake cues derived from the fusion of spatial and frequency features, enabling unified detection of live faces and all categories of attacks.
	Specifically, the attack-agnostic prompt module generates generic live and fake prompts within the language branch to extract corresponding generic representations from both live and fake faces, guiding the model to learn a unified feature space for unified attack detection. Meanwhile, the module adaptively generates the live/fake conditional bias from the original spatial and frequency information to optimize the generic prompts accordingly, reducing the impact of intra-class variations.
	We further propose a dual-stream cues fusion framework in the vision branch, which leverages frequency information to complement subtle cues that are difficult to capture in the spatial domain. In addition, a frequency compression block is utilized in the frequency stream, which reduces redundancy in frequency features while preserving the diversity of crucial cues. We also establish new challenging protocols to facilitate unified face attack detection effectiveness.
	Experimental results demonstrate that the proposed method significantly improves performance in detecting physical and digital face attacks, achieving state-of-the-art results.
\end{abstract}
\begin{IEEEkeywords}
	Face Anti-Spoofing, Deepfake Detection, Unified Feature Space, Frequency-Aware, Cues Fusion.
\end{IEEEkeywords}

\section{Introduction}
\IEEEPARstart{F}{ace} Attack Detection is a challenging task aiming to simultaneously detect facial physical attacks (PAs) and digital attacks (DAs). This technology is critical in security verification scenarios, such as smartphone unlocking, access control, and secure transactions. Previous research on face attack detection focuses on a single type of attack, such as physical attack detection (PAD) \cite{liu2018learning,george2020learning,yu2020fas,liu2021face,cai2022learning,guo2022multi,liu2022source,cai2024towards,liu2024cfpl} and digital attack detection (DAD) \cite{tolosana2020deepfakes,liu2023making,zhao2021multi}.
\IEEEpubidadjcol

\begin{figure}[!t]
	\centering
	\includegraphics[width=1.0\linewidth]{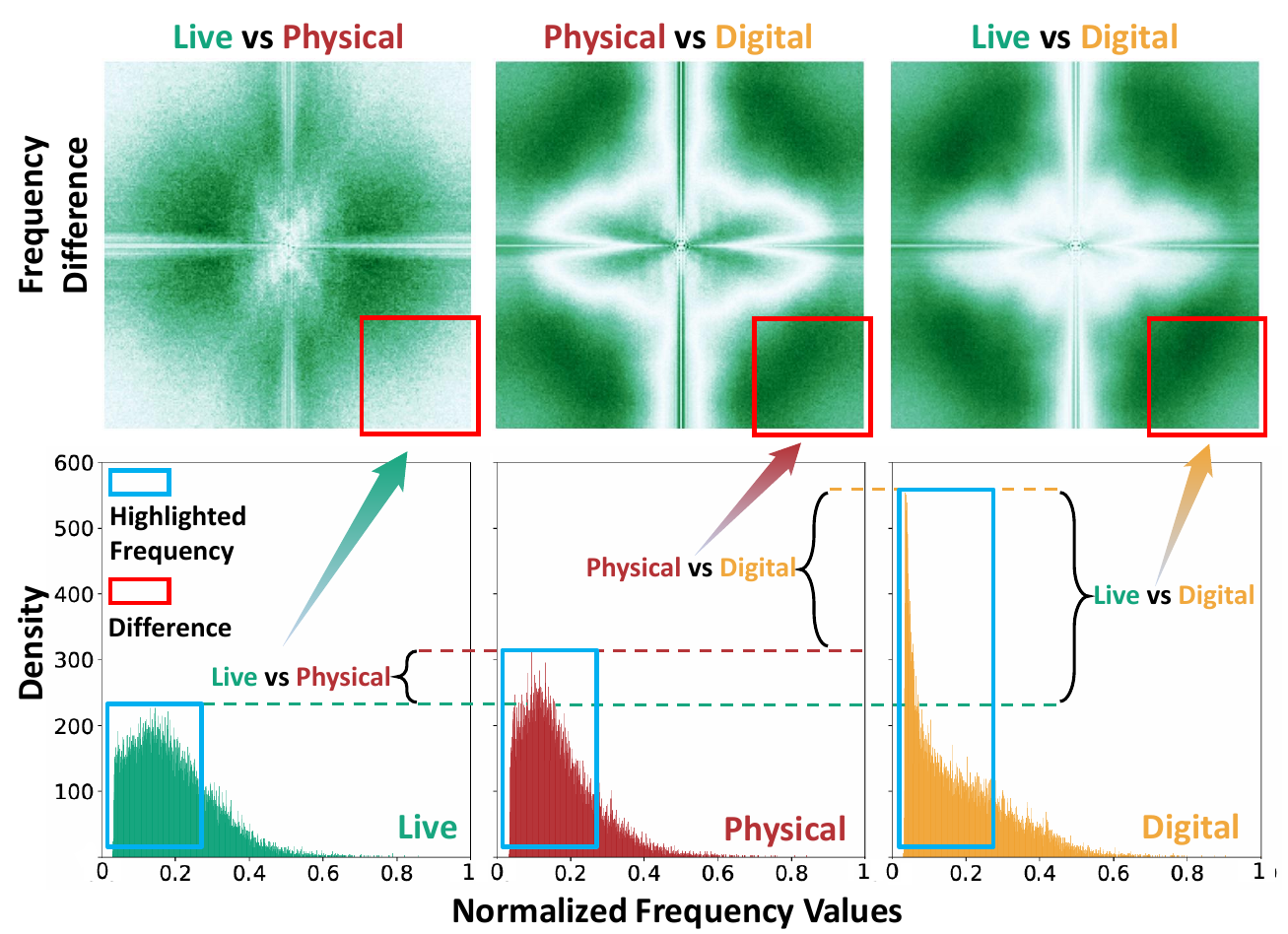}
	\caption{The figure illustrates the frequency differences among live faces, physical attacks (PAs), and digital attacks (DAs) in the UniAttackData \cite{fang2024unified}. The frequency density histograms are computed from Fourier Transform maps, averaging over 1,000 samples per category. Notably, higher frequency corresponds to a lower value, and the blue frame highlights the significant frequency differences among the three categories. To further examine these differences, we visualize the frequency difference maps, in which higher frequency components are concentrated at the periphery, and lighter colors indicate more minor differences between the compared categories. The results demonstrate substantial variations in frequency information across the three categories, suggesting that frequency features can be recognized as valuable indicators for unified attack detection.}
	\label{figure1}
\end{figure}

In recent years, advancements in PAD \cite{george2019biometric,george2021cross,zheng2024mfae,le2024grad,zhou2023instance,cai2024s,zheng2024famim} shift from handcrafted feature-based methods with limited attack handling capabilities to deep learning approaches \cite{zhang2020face,yu2020searching}. Initially, binary classification models \cite{wang2021multi} are widely adopted, but their poor generalization ability prompte the introduction of depth-based methods \cite{wu2021dual}, which capture the geometric structure of live faces. Subsequently, multi-modal and flexible modality approaches \cite{lin2024suppress} are developed by integrating multi-spectral signals to enhance robustness. However, these methods still struggle with cross-domain adaptation \cite{wang2020unsupervised}, leading to the emergence of domain generalization \cite{liu2021dual,qin2021meta,shao2020regularized,liu2023towards,liu2024cfpl}.

Compared to PAD, traditional DAD approaches \cite{dang2020detection,li2020face,rossler2019faceforensics++} typically employ CNN architectures to extract deep features for distinguishing live from fake samples. Despite their initial success, these methods demonstrate limited generalizability when confronted with advanced deepfake technologies. This limitation spurs the development of disentangled representation \cite{liu2021heterogeneous,li2022artifacts,yan2023ucf} and self-supervised learning \cite{chen2022self,haliassos2022leveraging}, aiming to reduce dependency on annotated datasets. Nevertheless, intrinsic attack patterns are still not comprehensively captured by existing methods, resulting in the development of multi-cue analysis methods that focus on specific forgery indicators, such as noise statistics \cite{gu2022exploiting,fei2022learning} and frequency characteristics \cite{luo2021generalizing,liu2021spatial}.

Despite the promise of these improvements, most existing methods are specifically tailored for detecting certain categories of attacks. Due to significant intra-class variations among different attacks(illustrated in Fig.~\ref{figure3}), models struggle to capture generic live and fake representations applicable across all categories. This limitation motivates our exploration of unified attack detection (UAD) within a unified feature space. Current UAD approaches \cite{deb2023unified,fang2024unified,zou2024softmoe,yu2024benchmarking} predominantly focus on spatial features, with only a subset of methods \cite{cao2024towards} incorporating frequency information. However, neither approach fully exploits the potential of frequency cues. Through frequency domain analysis of UniAttackData \cite{fang2024unified} (illustrated in Fig.~\ref{figure1}), we observe distinct distributions of frequency features across different facial categories. This further motivates us to explore spatial and frequency cues for more robust and generalizable attack detection.

To address the aforementioned challenges, we propose a novel frequency-aware and attack-agnostic CLIP model termed FA\textsuperscript{3}-CLIP, which constructs a unified feature space to detect all facial categories. Specifically, the attack-agnostic prompt module generates generic live and fake prompts in the language branch, enabling extraction of corresponding generic representations from both live and fake faces, regardless of the specific type. Additionally, live and fake bias generators dynamically optimize these prompts by highlighting discriminative cues within live and fake face classes to mitigate intra-class variations. We also employ the normalized temperature-scaled cross entropy Loss to prevent overlapping cues between bias generators. In the vision branch, we introduce a dual-stream cues fusion framework to extract more comprehensive cues. Considering the nature of deep features across different layers of the Vision Transformer, we design a layer-wise frequency generator to aggregate frequency features across multiple layers, thereby enriching the feature representation and effectively complementing spatial information. Furthermore, given the potential redundancy across multiple layers, a frequency compression module is incorporated to efficiently reduce redundant frequency information while preserving crucial cue diversity. As performance approaches saturation, we establish new challenging protocols to evaluate the effectiveness of UAD.

Our main contributions are summarized as follows:
\begin{itemize}
	\item{An attack-agnostic prompt learning is introduced to characterize the generic live and fake representations for various categories of facial images.}
	\item{We design a dual-stream cues fusion framework to leverage the multilayer frequency information to complement the spatial features, and utilize a frequency compression block to compact and refine frequency cues.}
	\item{Challenging evaluation protocols are established to rigorously evaluate unified face attack detection methods.}
	\item{Extensive experiments on the unified attack datasets demonstrate that the proposed method significantly improves performance, achieving state-of-the-art results.}
\end{itemize}

\section{Related Work}
\subsection{Face Anti-Spoofing}
Face anti-spoofing (FAS) is critical for securing face recognition systems by distinguishing live faces from presentation attacks (e.g., prints, replays, 3D masks~\cite{liu2022contrastive}). 
With the advancement of deep learning, researchers explore CNN to formulate FAS as a binary classification problem \cite{wang2021multi,liu2021casia,liu2022disentangling}. To further exploit intrinsic spoof patterns, pixel-wise supervision based on depth maps \cite{george2019deep}, reflection maps \cite{yu2020searching}, and texture maps \cite{zhang2020face} is introduced to guide feature learning at a finer granularity. However, the presence of unseen attacks and domain shifts can significantly degrade the performance of FAS systems. To address this limitation, domain adaptation \cite{wang2020unsupervised,liu2022source,wang2021self} and domain generalization \cite{wang2020cross,jia2020single,zheng2024learning,wang2024tf,liu2024moeit} are proposed to enhance robustness across attack variations. Apart from these mainstream approaches, researchers also investigate adversarial learning \cite{jia2020single}, meta learning \cite{cai2022learning} and continual learning \cite{guo2022multi,rostami2021detection,cai2023rehearsal} to handle novel or unexpected spoof scenarios. Recently, some flexible modal based FAS algorithms~\cite{george2021cross,ijcai2022p165,Liu2023FMViTFM,liu2024fm} have also become increasingly popular.

\subsection{Face Forgery Detection}
Face forgery detection (FFD) aims to discriminate live faces from digitally manipulated images \cite{liu2023making,yang2021mtd,sun2022dual,CLIP-forgery-detection}. Early approaches typically employ CNN-based backbones to perform binary classification on cropped facial images \cite{nguyen2019capsule,rossler2019faceforensics++}. With the increasing photorealism of forged faces, these generic methods struggle to capture subtle artifacts inherent in manipulated images. This challenge motivates recent advances in disentangled representation \cite{li2022artifacts,yan2023ucf,yan2023ucf} and self-supervised learning \cite{ chen2022self,haliassos2022leveraging,chen2022self}, enhancing generalization against unseen manipulation techniques. Additionally, researchers focus on exploiting more specialized fake cues, such as noise statistics, local texture inconsistencies, and frequency signatures. For example, Zhao et al. \cite{zhao2021multi} propose a texture enhancement module that efficiently aggregates texture information and high-level semantic features from multiple local regions. Qian et al. \cite{qian2020thinking} and Wang et al. \cite{wang2023dynamic} utilize frequency-aware models to uncover hidden spectral artifacts. Fei et al. \cite{fei2022learning} specifically leverage subtle noise patterns and visual artifacts to strengthen the representation of fake cues.

\subsection{Physical-Digital Attack Detection}
Previous FAS and FFD focus on specific attack types without considering the interaction between PAs and DAs, leading to the proposal of UAD as a solution to handle all attacks simultaneously. However, due to the significant intra-class variations across different attacks, learning a unified feature space that captures generic cues for distinguishing live faces from all fake images remains challenging. To alleviate these variations, Deb et al. \cite{deb2023unified} employ a clustering method to organize similar attack types and adopt a multi-task learning framework to distinguish both unique traits of live and fake cues. Similarly, Fang et al. \cite{fang2024unified} propose a partition of unified ID face dataset, aligning live and fake samples (both PAs and DAs) under the same identity, thus allowing models to gain more comprehensive insights from the differences between live and fake samples. Beyond better data organization, recent work also explores data augmentation and mixture-of-experts mechanisms. For example, Zou et al. \cite{zou2024softmoe} integrate a SoftMoE module into the CLIP framework to efficiently handle sparse feature distributions. Likewise, He et al. \cite{he2024joint} augment live samples by simulating fake cues of both PAs and DAs, substantially improving the model’s capacity to detect unseen attack types. Furthermore, some researches incorporate additional discriminative cues beyond spatial features. Yu et al. \cite{yu2024benchmarking} introduce a dual-branch physiological network that exploits spatiotemporal rPPG signal maps and continuous wavelet transforms to enhance periodic discrimination, enabling unified detection through both visual appearance and physiological signals.  Cao et al. \cite{cao2024towards} highlight the disparities of live and fake faces by extracting compact representations of live samples from spatial and frequency features, then reconstructing distributions of live faces to reveal crucial differences and discard irrelevant information.

Most UAD methods predominantly rely on the visual modality, with only a few studies adopting contrastive loss to learn a unified feature space for textual prompts and visual features. However, these approaches have yet to fully exploit the relationship between visual and textual modalities. In response to this gap, we propose attack-agnostic prompt learning that adaptively generate generic live and fake prompts to capture the distinctions between live faces and all categories of attack. Furthermore, we introduce a dual-stream cues fusion framework that integrates frequency information and spatial features, enabling the extraction of more comprehensive and discriminative cues for unified face attack detection.

\section{Methodology}

\subsection{Preliminary}
\noindent{\bf{Contrastive Language-Image Pre-training (CLIP).}} CLIP \cite{radford2021learning} is a pre-trained vision-language model designed to project an image $\boldsymbol{I} \in \mathbb{R}^{H \times W \times 3}$ and a corresponding text prompt template $\boldsymbol{T}$ into a unified feature space to predict the correct image-text pairs during training. After training, CLIP can be further used for zero-shot recognition. 

The CLIP consists of an image encoder $\mathcal{V}(\cdot)$ and a text encoder $\mathcal{T}(\cdot)$. In the vision branch, the image $\boldsymbol{I}$ is divided into $n=196$ patches, and projected into patch tokens $\boldsymbol{E} = \{ \boldsymbol{e}_i \}_{i=1}^{n} \in \mathbb{R}^{n \times d_{v}}$ where $d_{v} = 768$ is the dimension of the feature. And the input sequence $\boldsymbol{Z}_x = \{\texttt{CLS}, \boldsymbol{e}_1, \boldsymbol{e}_2, \ldots, \boldsymbol{e}_n\} \in \mathbb{R}^{(1+n) \times d_{v}}$ is constructed by a learnable class token $\texttt{[CLS]}$ and patch tokens $\boldsymbol{E}$. After processing by the vision transformer, the output class token from the final layer is denoted as the visual feature $\boldsymbol{x} = \mathcal{V}(\boldsymbol{I}) \in \mathbb{R}^{d_{v}}$. In the language branch, the text template $\boldsymbol{T}$ typically uses a fixed context of $M$ words, such as ``a photo of \texttt{<CLASS>} face.'', along with the tokens $\texttt{[SOS]}$ and $\texttt{[EOS]}$ to construct a text sequence $\boldsymbol{Z}_t = \{\boldsymbol{t}_{\texttt{SOS}}, \boldsymbol{t}_1, \boldsymbol{t}_2, \ldots, \boldsymbol{t}_M, c_k, \boldsymbol{t}_{\texttt{EOS}}\} \in \mathbb{R}^{77 \times d_{t}}$, where $d_t = 512$ is the dimension of the textual embedding and $c_k$ is the $k$-th label. The text feature is obtained by the $\texttt{[EOS]}$ token in last transformer layer as $\boldsymbol{t} = \mathcal{T}(\boldsymbol{T}) \in \mathbb{R}^{K \times d_t}$ for $K$ categories. Both the visual feature $\boldsymbol{x}$ and the text feature $\boldsymbol{t}$ are projected by a linear layer into a unified feature space with dimension $d_{vt} = 512$. 

During zero-shot inference, the text encoder generates a weight vector set $\{\boldsymbol{w}_k\}_{k=1}^K$, with each vector corresponding to a category to calculate the similarity to the visual feature. The probability of assigning the visual feature $\boldsymbol{x}$ to the category $k$ is given by:
\begin{equation}
	\label{eq:1}
	p(\boldsymbol{y} = k \mid \boldsymbol{x}) = \frac{\exp\bigl(\mathrm{sim}(\boldsymbol{x}, \boldsymbol{w}_k)/\tau\bigr)}
	{\sum_{i=1}^{K} \exp\bigl(\mathrm{sim}(\boldsymbol{x}, \boldsymbol{w}_i)/\tau\bigr)},
\end{equation}
where $\mathrm{sim}(\cdot,\cdot)$ denotes a similarity function and $\tau$ is a temperature parameter.

\noindent{\bf{Context Optimization (CoOp).}} CoOp \cite{zhou2022learning} adopts learnable context vectors instead of a fixed prompt to facilitate CLIP’s adaptation to downstream tasks. These learnable context vectors are defined as $\{\boldsymbol{v}_1, \boldsymbol{v}_2, \ldots, \boldsymbol{v}_M\}$, and each vector has the same dimension. In the language branch, the prompt is represented as $\boldsymbol{Z}_k = \bigl\{ \boldsymbol{v}_{\text{[SOS]}}, \boldsymbol{v}_1, \boldsymbol{v}_2, \ldots, \boldsymbol{v}_M, c_k, \boldsymbol{v}_{\text{[EOS]}} \bigr\} \in \mathbb{R}^{77 \times d_t}$ where $c_k$ is the $k$-th label. Despite alleviating constraints imposed by fixed templates, CoOp still exhibits limited generalizability to unseen classes.

\noindent{\textbf{Conditional Context Optimization (CoCoOp).}} To address this issue, CoCoOp \cite{zhou2022conditional} introduces conditional context to adapt to instance-level inputs. For each vector in $\boldsymbol{Z}_k$, CoCoOp utilizes a Meta-Net $h(\cdot)$ to generate a bias from the visual feature to optimize the context as $\boldsymbol{v}_m(\boldsymbol{x}) = \boldsymbol{v}_m + \boldsymbol{\pi}$, where $\boldsymbol{\pi} = h(\boldsymbol{x})$. Hence, the conditional prompt for the $k$-th label is given by $\boldsymbol{Z}_k(\boldsymbol{x}) = \bigl\{ \boldsymbol{v}_{\text{[SOS]}}, \boldsymbol{v}_1(\boldsymbol{x}), \boldsymbol{v}_2(\boldsymbol{x}), \ldots, \boldsymbol{v}_M(\boldsymbol{x}), c_k, \boldsymbol{v}_{\text{[EOS]}} \bigr\} \in \mathbb{R}^{77 \times d_t}$, and the text features for all $K$ labels are then obtained by $\mathbf{\boldsymbol{t}} = \mathcal{T}\bigl(\boldsymbol{Z}_k(\boldsymbol{x})\bigr) \in \mathbb{R}^{K \times d_t}$.

\begin{figure*}[htbp]
	\centering
	\includegraphics[width=\linewidth]{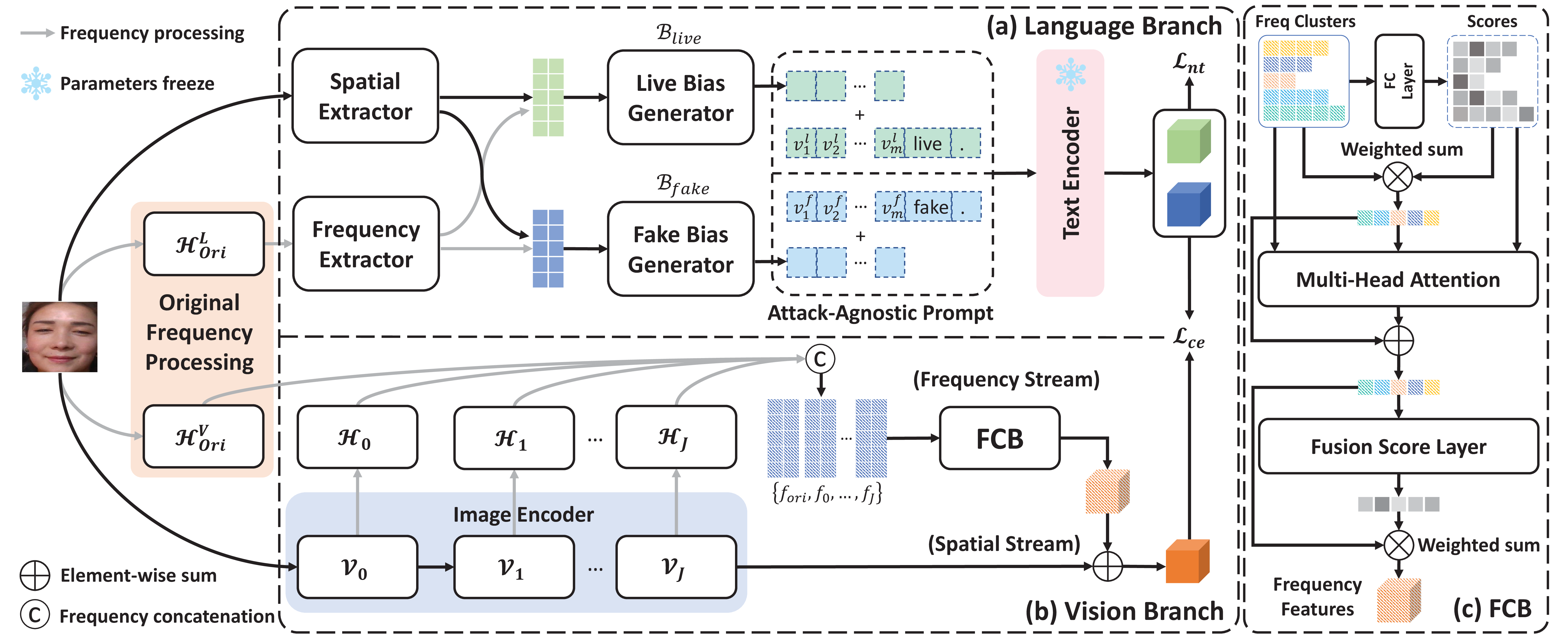}
	\caption{The architecture of the attack-agnostic prompt learning and dual-stream cues fusion framework in FA\textsuperscript{3}-CLIP. The frequency generators $\mathcal{H}_{Ori}^{V}$ and $\mathcal{H}_{Ori}^{L}$ are introduced to extract the frequency information from  the original image in vision and language branches. In vision branch, the frequency feature at each vision transformer layer is denoted by $\boldsymbol{f}_j=\mathcal{H}_j(\mathcal{V}_j(\boldsymbol{Z}_j))$ where $\mathcal{V}_j(\cdot)$ denotes the $j$-th vision transformer layer, the $\boldsymbol{Z}_j$ represents its corresponding input tokens. Then the multi-layer frequency features are compressed through the frequency compression block (FCB) and integrated with visual features. In language branch, the bias generators $\mathcal{B}_i(\cdot)$ are employed to optimize the generic live and fake prompts. Additionally, FA\textsuperscript{3}-CLIP incorporates constraints based on both normalized temperature-scaled cross-entropy $\mathcal{L}_{nt}$ and standard cross-entropy $\mathcal{L}_{ce}$.} 
	\label{figure2}
\end{figure*}

\subsection{Overview}
This proposes Frequency-Aware and Attack-Agnostic CLIP (FA\textsuperscript{3}-CLIP), which employs an attack-agnostic prompt generator to generate generic live and fake prompts for all facial categories. Considering that attack methods produce different fake cues in spatial and frequency domains, we further incorporate frequency information to assist the model in uncovering the essential difference between live and fake traces. As depicted in Fig.~\ref{figure2}, we introduce two conditional bias generators to optimize the generic live and fake contexts in the language branch, resulting in more discriminative live and fake representations for each image. To comprehensively capture live and fake traces within the vision branch and align them with the generalized embeddings, we introduce a dual-stream cues fusion framework, which integrates frequency cues from various vision Transformer layers to complement spatial features. To minimize redundant frequency information without losing diversity, we compress the patches with similar frequency characteristics while preserving multi-level frequency information. 

\subsection{Attack-Agnostic Prompt Learning}
The language branch consists of three main operations, frequency features generation, spatial-frequency fusion, and generic prompts generation. As shown in Fig.~\ref{figure1}, live and fake faces exhibit substantial differences in frequency components, motivating us to complement spatial features with frequency cues.

\noindent
\textbf{Frequency Features Generation (FFG).}
As high-frequency components are predominantly concentrated around the periphery of the frequency spectrum, and differences among all face categories are mainly observed in higher frequency regions, a high-pass filter is employed to isolate the higher frequency components. This process compels the generator to focus on frequency difference regions between live and fake faces, ensuring that the model consistently prioritizes discriminative frequency features. Specifically, the image is first transformed into the frequency domain $\mathcal{F}_H,_W(\boldsymbol{I}) \in \mathbb{R}^{H \times W \times 3}$, where $\mathcal{F}$ denotes the Fast Fourier Transform (FFT). Subsequently, we extract the frequency map $\boldsymbol{f}_h= \in \mathbb{R}^{H \times W \times 3}$ using a high-pass filter $\mathcal{M}_h \in \mathbb{R}^{H \times W}$:
\begin{equation}
	\begin{aligned}
		&\boldsymbol{f}_h = \mathcal{IF}_{H,W}(\mathcal{M}_h(\mathcal{F}_H,_W(\boldsymbol{I})), \\
		&\mathcal{M}_h(\boldsymbol{f}_{i,j}) = \begin{cases} 
			\boldsymbol{f}_{i,j}, & otherwise, \\
			0, & if |i| < \alpha W, |j| < \alpha H,
		\end{cases}
	\end{aligned}
\end{equation}
where the origin of the frequency domain is placed at the center of the image, and $\mathcal{IF}$ denotes the inverse FFT (iFFT).

To effectively extract frequency information, CNNs are separately applied on the amplitude and phase spectra, enhancing the discriminative capability of the frequency generator. The process is formulated as:
\begin{equation}
	\begin{aligned}
		&\boldsymbol{f} = \boldsymbol{f}_{am} + \boldsymbol{f}_{ph} \text{i} = \mathcal{F}_{H,W}(\boldsymbol{f}_h), \\
		&\widetilde{\boldsymbol{f}_{am}} = \phi_1(\boldsymbol{f}_{am}), \\
		&\widetilde{\boldsymbol{f}_{ph}} = \phi_2(\boldsymbol{f}_{ph}), \\
		&\boldsymbol{I}_h = \mathcal{IF}_{H,W}(\widetilde{\boldsymbol{f}_{am}} + \widetilde{\boldsymbol{f}_{ph}} \text{i}),
	\end{aligned}
\end{equation}
where $\boldsymbol{f}_{am}$ and $\boldsymbol{f}_{ph}$ denote amplitude spectrum and phase spectrum of frequency map $\boldsymbol{f}_h$, $\phi_1$ and $\phi_2$ represent different CNN blocks, and $\boldsymbol{I}_h \in \mathbb{R}^{H \times W \times 3}$ indicates the extracted frequency feature. The entire extraction process of FFG in language branch is denoted as $\boldsymbol{I}_h=\mathcal{H}_{Ori}^{L}(\boldsymbol{I})$.

\noindent
\textbf{Spatial-Frequency Fusion in Language Branch.}
To project the input spatial and frequency features into a unified feature space, We adopt two separate feature extractors to process the original features from $\boldsymbol{I}$ and frequency features from $\boldsymbol{I}_h$, respectively. Then extracted features are combined using a learnable parameter $\beta$ to adaptively balance their contributions. Formally, the process is defined as $\boldsymbol{I}_f=\beta\,\psi_1(\boldsymbol{I})+(1 - \beta)\,\psi_2(\boldsymbol{I}_h)$ where $\psi_1$ and $\psi_2$ indicate different CNN blocks, and $\boldsymbol{I}_f\in \mathbb{R}^{n \times d_{t}}$ denotes the fused input feature for the original image in language branch.

\noindent
\textbf{Generic Prompts Generation.}
The unified attack detection task is designed to enable the model to separate live and fake faces within a unified feature space, classifying both physical and digital attacks as fake. In line with this objective, we propose attack-agnostic prompt learning, which constructs learnable generic live and fake prompts for each image to facilitate the learning of a more discriminative textual feature space. The learnable generic context vectors are defined as $\boldsymbol{P}_k = \{ \boldsymbol{v}_\texttt{[SOS]}, \boldsymbol{v}_1^k, \boldsymbol{v}_2^k, \dots, \boldsymbol{v}_M^k, c_k, \boldsymbol{v}_\texttt{[EOS]} \} \in \mathbb{R}^{77 \times d_{t}}$, where $k \in \{\texttt{live}, \texttt{fake}\}$ indicates the class label. Distinct from CoCoOp \cite{zhou2022conditional}, two independent bias generators $\mathcal{B}_k(\cdot)$ are introduced to optimize the context vectors as $\boldsymbol{v}_m^k (\boldsymbol{I}_f) = \boldsymbol{v}_m^k + \boldsymbol{b}_k$, where $\boldsymbol{b}_k = \mathcal{B}_k (\boldsymbol{I}_f)$. The attack-agnostic prompt is describe as $\boldsymbol{P}_k(\boldsymbol{I}_f) = \{ \boldsymbol{v}_\texttt{[SOS]}, \boldsymbol{v}_1^k(\boldsymbol{I}_f), \boldsymbol{v}_2^k(\boldsymbol{I}_f), \dots, \boldsymbol{v}_M^k(\boldsymbol{I}_f), c_k, \boldsymbol{v}_\texttt{[EOS]} \} \in \mathbb{R}^{77 \times d_{t}}$, which incorporates the instance-conditional information from the fused input feature $\boldsymbol{I}_f$. Subsequently, the text feature $\mathbf{\boldsymbol{t}}_k$ is produced by $\mathbf{\boldsymbol{t}}_k = \mathcal{T}\bigl(\boldsymbol{P}_k(\boldsymbol{I}_f)\bigr) \in \mathbb{R}^{K \times d_t}$.

While CoCoOp improves generalization via instance-conditioned contexts, its unified context may inadvertently blend live and fake cues, leading to overlapping representations within textual feature space. This overlap limits discriminability, especially in tasks requiring strict separation between live and fake cues. The separate contexts adopted in our approach ensure that live representations emphasize live cues, while fake representations focus on attack information from both PAs and DAs, thereby enhancing the model’s ability to disentangle live and fake features.

To ensure compact textual embeddings for live samples while clustering all fake samples into a fake class and distinguishing live and fake representations, we adopt a Normalized Temperature-scaled Cross-Entropy loss, defined as:
\begin{equation}
	\mathcal{L}_{nt} = -\frac{1}{N} \sum_{i=1}^N \log \frac{\exp\left(\boldsymbol{t}_k^\top \boldsymbol{t}_i / \tau\right)}{\sum_{j \in \{\texttt{live}, \texttt{fake}\}} \exp\left(\boldsymbol{t}_k^\top \boldsymbol{t}_j / \tau\right)},
\end{equation}
where $N$ denotes the number of samples in the current batch, $\boldsymbol{t}_i$ represents the text feature of the $i$-th sample, $\boldsymbol{t}_k$ is the text prototype for class $k \in \{\texttt{live}, \texttt{fake}\}$, and $\tau$ is a temperature parameter.

\subsection{Dual-Stream Cues Fusion Framework}
\label{DSCFF_subsection}
Due to the neglect of frequency cues, relying solely on the \texttt{CLS} token from the final transformer layer is insufficient for obtaining a comprehensive visual representation. To mitigate this issue, we propose extracting frequency features from each vision transformer layer, preserving a broader spectrum of cues while retaining the original frequency information. The process of each layer is described as $\boldsymbol{f}_i=\mathcal{H}_i(\boldsymbol{x}_i)\in \mathbb{R}^{n \times d_v}$ where $\boldsymbol{x}_i$ represents the output tokens of the $i$-th transformer layer, excluding the \texttt{CLS} token. Frequency features $\boldsymbol{f}_i$ from each layer are concatenated to form the final representation $\boldsymbol{Z}_h = \{\boldsymbol{f}_{Ori}, \boldsymbol{f}_0, \boldsymbol{f}_1, \ldots, \boldsymbol{f}_J\} \in \mathbb{R}^{(1+J) \times n \times d_{v}}$ where $\boldsymbol{f}_{Ori}=\mathcal{H}_{Ori}^V(\boldsymbol{I})$ denotes the original frequency features, and the output imensions of $\mathcal{H}_{Ori}^V$ (vision branch) and $\mathcal{H}_{Ori}^L$ (language branch) differ.

\noindent
\textbf{Frequency Compression Block (FCB).} The frequency extraction network produces a substantial volume of tokens, a portion of which exhibit mirrored characteristics (illustrated in Fig.~\ref{figure6}). Directly processing $\boldsymbol{Z}_h$ through the Transformer during final cue extraction may cause quantitatively dominant tokens with similar characteristics to overshadow minority tokens containing unique information. Therefore, we utilize a dynamic features compression mechanism to preserve cue diversity while balancing their contributions.

Inspired by \cite{zhu2023umiformer}, we establish compression relationships among similar tokens across layers. The tokens in $\boldsymbol{Z}_h$ are grouped into $d$ clusters using KNN-based clustering, where each cluster represents a set of similar cues. Further, we employ a fully connected layer as scoring layer to assign weights to each token, enabling intra-cluster weighted summation. This process yields compressed tokens $\boldsymbol{Z}_h'\in \mathbb{R}^{d \times d_v}$, effectively balancing diversity and representativeness. It can be formulated as:
\begin{equation}
	\boldsymbol{Z}_h' = \sum_{i=1}^{d} \boldsymbol{s}_i \cdot \boldsymbol{D}_i,
\end{equation}
where $\boldsymbol{D}_i$ denotes the $i$-th cluster of tokens, $\boldsymbol{s}_i$ represents intra-cluster weights assigned by the scoring layer, and $d$ is the total number of clusters. We project compressed frequency features into the visual feature space to align with spatial features $\boldsymbol{x}$ using Transformer attention as follows:
\begin{equation}
	\boldsymbol{x}_h = \varphi \left(\mathrm{softmax} \left( \frac{Q (\boldsymbol{Z}_h') K^\top (\boldsymbol{Z}_h)}{\sqrt{d_k}} + S \right) V (\boldsymbol{Z}_h) \right),
\end{equation}
where $S$ denotes the weights of $\boldsymbol{Z}_h$, $\boldsymbol{x}_h \in \mathbb{R}^{d_v}$ represents the frequency feature, and $\varphi$ indicates a fusion score layer (consists of MLP) in visual branch. Finally, the comprehensive visual feature is described as $\boldsymbol{x}_f=\boldsymbol{x}+(1 - \gamma)\,\boldsymbol{x}_h \in \mathbb{R}^{d_v}$ where $\gamma$ is a learnable parameter balancing spatial and frequency features.

\subsection{Total Loss Function}
We employ a cross-entropy loss function $\mathcal{C}(\cdot,\cdot)$ to quantify discrepancy between predicted labels and ground truths in classification dataset $\mathcal{D}$. The parameters of the text encoder $\mathcal{T}(\cdot)$ remain frozen while the other parameters are optimized using the loss function:
\begin{equation}
	\mathcal{L}_{ce} = \arg \min_{\theta \notin \mathcal{T}} \mathbb{E}_{(\boldsymbol{I}, \boldsymbol{y}) \in \mathcal{D}} \mathcal{C}(\boldsymbol{y}, \texttt{sim}(\boldsymbol{x}_f, \boldsymbol{t}_k)),
\end{equation}
where $\boldsymbol{t}_k$ represents the text feature and $\boldsymbol{x}_f$ denotes the visual feature. We combine all the cross-entropy loss functions to form the final loss, as expressed by the following equation:
\begin{equation}
	\mathcal{L}_{total} = \mathcal{L}_{nt} + \mathcal{L}_{ce}.
\end{equation}

\subsection{Protocols}
Existing protocols for Digital and Physical Attack Detection usually involve dataset splits resulting in substantial overlap or high similarity among the training, validation, and test sets, particularly for digital attacks. In \textit{protocol 1} of UniAttackData \cite{fang2024unified} (as shown in Table~\ref{tab1}), the dataset comprises 1800 unique identities, each associated with live face, physical attack, and digital attack samples. For both the live and physical attack categories, 600 identities are allocated for training, 300 for validation, and 900 for testing, ensuring no identity overlaps among the these sets. However, for digital attack, all 1800 identities are utilized across training, validation, and test sets, leading to potential identity information leakage. Additionally, certain subtypes of digital attacks exhibit high intra-category similarity (illustrated in Fig.~\ref{figure3}), further exacerbating this issue by limiting the diversity of fake samples during training and evaluation.

To address these limitations, we redefine the experimental protocols based on the following guidelines to guarantee strict independence among training, validation, and test sets:

\subsubsection{Strict Non-Overlapping ID Assignment} The identities in the training, validation, and test sets are partitioned explicitly to ensure complete exclusivity, preventing any overlap.

\subsubsection{Strategic Category Distribution} We enhance identity independence by strategically distributing attack subtypes with high intra-category similarity across different data splits. This ensures a more realistic evaluation of generalization capability across diverse attack patterns.

\subsubsection{Balanced Distribution of Attack Types} To ensure fairness and robustness, we balance the proportions of live, physical, and digital attacks in each split. This approach mitigates the risk of overfitting to specific attack types.

As shown in Table~\ref{tab1}, we propose three distinct protocols. The \textit{Protocol 1.1} and \textit{Protocol 1.2} provide independent evaluations of the "digital" category, specifically focusing on advanced and deepfake subtypes. In these protocols, the live face and physical attack categories follow the original distribution (600 for training, 300 for validation, and 900 for testing). In \textit{Protocol 1.1}, the advanced subtype follows the standard data split, while the deepfake subtype is excluded from both the training and validation sets and evaluated only on a distinct identity set during testing. Conversely, in \textit{Protocol 1.2}, the advanced subtype is similarly excluded from training and validation, with evaluation conducted solely on the testing set. Finally, \textit{Protocol 1.3} retains the standard distribution for all digital subtypes. These protocols are tailored explicitly for unified face attack detection tasks, simultaneously evaluating the robustness against both physical and digital attack detections.

\begin{table*}
	\caption{The new Protocols for UniAttackData \cite{fang2024unified} enforce strict non-overlapping ID assignment. The IDs of digital attack in \textit{Protocol 1} are not aligned with live face and physical attack that lead to potential ID leakage. New Protocols aims to prevent ID leakage and ensure a more rigorous evaluation. The number of IDs is highlighted in the gray columns.}
	\begin{tabular}{|c|c|cccccccc|c|}
		\hline
		\multirow{3}{*}{Protocol} & \multirow{3}{*}{Class} & \multicolumn{8}{c|}{Types}                                                                                                                                                                                                                                                             & \multirow{2}{*}{\# Total}    \\ \cline{3-10} 
		&                        & \multicolumn{2}{c|}{\# Live}                                     & \multicolumn{2}{c|}{\# Physical}                                     & \multicolumn{2}{c|}{\# Advanced}                                      & \multicolumn{2}{c|}{\# DeepFake}                                                               &                               \\ \cline{3-11}

		&                        & \multicolumn{1}{c|}{\# Img Num} & \multicolumn{1}{c|}{\cellcolor{gray!30}\# ID Num} & \multicolumn{1}{c|}{\# Img Num} & \multicolumn{1}{c|}{\cellcolor{gray!30}\# ID Num} & \multicolumn{1}{c|}{\# Img Num} & \multicolumn{1}{c|}{\cellcolor{gray!30}\# ID Num} & \multicolumn{1}{c|}{\# Img Num} & \cellcolor{gray!30}\# ID Num & \# Img Num \\ \hline
		\multirow{3}{*}{P1}       & trian                  & \multicolumn{1}{c|}{3000}       & \multicolumn{1}{c|}{\cellcolor{gray!30}600}       & \multicolumn{1}{c|}{1800}       & \multicolumn{1}{c|}{\cellcolor{gray!30}600}       & \multicolumn{1}{c|}{1800}       & \multicolumn{1}{c|}{\cellcolor{gray!30}1800}      & \multicolumn{1}{c|}{1800}       & \cellcolor{gray!30}1800      & 8400       \\ 
		& eval                   & \multicolumn{1}{c|}{1500}       & \multicolumn{1}{c|}{\cellcolor{gray!30}300}       & \multicolumn{1}{c|}{900}        & \multicolumn{1}{c|}{\cellcolor{gray!30}300}       & \multicolumn{1}{c|}{1800}       & \multicolumn{1}{c|}{\cellcolor{gray!30}1800}      & \multicolumn{1}{c|}{1800}       & \cellcolor{gray!30}1800      & 6000       \\ 
		& test                   & \multicolumn{1}{c|}{4500}       & \multicolumn{1}{c|}{\cellcolor{gray!30}900}       & \multicolumn{1}{c|}{2700}       & \multicolumn{1}{c|}{\cellcolor{gray!30}900}       & \multicolumn{1}{c|}{7106}       & \multicolumn{1}{c|}{\cellcolor{gray!30}1800}      & \multicolumn{1}{c|}{7200}       & \cellcolor{gray!30}1800      & 21506      \\ \hline
		\multirow{3}{*}{P1.1}     & trian                  & \multicolumn{1}{c|}{3000}       & \multicolumn{1}{c|}{\cellcolor{gray!30}600}       & \multicolumn{1}{c|}{1800}       & \multicolumn{1}{c|}{\cellcolor{gray!30}600}       & \multicolumn{1}{c|}{1200}       & \multicolumn{1}{c|}{\cellcolor{gray!30}600}       & \multicolumn{1}{c|}{0}          & \cellcolor{gray!30}0         & 6000       \\ 
		& eval                   & \multicolumn{1}{c|}{1500}       & \multicolumn{1}{c|}{\cellcolor{gray!30}300}       & \multicolumn{1}{c|}{900}        & \multicolumn{1}{c|}{\cellcolor{gray!30}300}       & \multicolumn{1}{c|}{300}        & \multicolumn{1}{c|}{\cellcolor{gray!30}300}       & \multicolumn{1}{c|}{0}          & \cellcolor{gray!30}0         & 2700       \\ 
		& test                   & \multicolumn{1}{c|}{4500}       & \multicolumn{1}{c|}{\cellcolor{gray!30}900}       & \multicolumn{1}{c|}{2700}       & \multicolumn{1}{c|}{\cellcolor{gray!30}900}       & \multicolumn{1}{c|}{2606}       & \multicolumn{1}{c|}{\cellcolor{gray!30}900}       & \multicolumn{1}{c|}{5425}       & \cellcolor{gray!30}900       & 15231      \\ \hline
		\multirow{3}{*}{P1.2}     & trian                  & \multicolumn{1}{c|}{3000}       & \multicolumn{1}{c|}{\cellcolor{gray!30}600}       & \multicolumn{1}{c|}{1800}       & \multicolumn{1}{c|}{\cellcolor{gray!30}600}       & \multicolumn{1}{c|}{0}          & \multicolumn{1}{c|}{\cellcolor{gray!30}0}         & \multicolumn{1}{c|}{1198}       & \cellcolor{gray!30}600       & 5998       \\ 
		& eval                   & \multicolumn{1}{c|}{1500}       & \multicolumn{1}{c|}{\cellcolor{gray!30}300}       & \multicolumn{1}{c|}{900}        & \multicolumn{1}{c|}{\cellcolor{gray!30}300}       & \multicolumn{1}{c|}{0}          & \multicolumn{1}{c|}{\cellcolor{gray!30}0}         & \multicolumn{1}{c|}{300}        & \cellcolor{gray!30}300       & 2700       \\ 
		& test                   & \multicolumn{1}{c|}{4500}       & \multicolumn{1}{c|}{\cellcolor{gray!30}900}       & \multicolumn{1}{c|}{2700}       & \multicolumn{1}{c|}{\cellcolor{gray!30}900}       & \multicolumn{1}{c|}{5306}       & \multicolumn{1}{c|}{\cellcolor{gray!30}900}       & \multicolumn{1}{c|}{2725}       & \cellcolor{gray!30}900       & 15231      \\ \hline
		\multirow{3}{*}{P1.3}     & trian                  & \multicolumn{1}{c|}{3000}       & \multicolumn{1}{c|}{\cellcolor{gray!30}600}       & \multicolumn{1}{c|}{1800}       & \multicolumn{1}{c|}{\cellcolor{gray!30}600}       & \multicolumn{1}{c|}{600}        & \multicolumn{1}{c|}{\cellcolor{gray!30}600}       & \multicolumn{1}{c|}{600}        & \cellcolor{gray!30}600       & 6000       \\ 
		& eval                   & \multicolumn{1}{c|}{1500}       & \multicolumn{1}{c|}{\cellcolor{gray!30}300}       & \multicolumn{1}{c|}{900}        & \multicolumn{1}{c|}{\cellcolor{gray!30}300}       & \multicolumn{1}{c|}{300}        & \multicolumn{1}{c|}{\cellcolor{gray!30}300}       & \multicolumn{1}{c|}{300}        & \cellcolor{gray!30}300       & 3000       \\ 
		& test                   & \multicolumn{1}{c|}{4500}       & \multicolumn{1}{c|}{\cellcolor{gray!30}900}       & \multicolumn{1}{c|}{2700}       & \multicolumn{1}{c|}{\cellcolor{gray!30}900}       & \multicolumn{1}{c|}{3506}       & \multicolumn{1}{c|}{\cellcolor{gray!30}900}       & \multicolumn{1}{c|}{3625}       & \cellcolor{gray!30}900       & 14331      \\ \hline
	\end{tabular}
	\label{tab1}
\end{table*}

\begin{figure}[!t]
	\centering
	\includegraphics[width=1.0\linewidth]{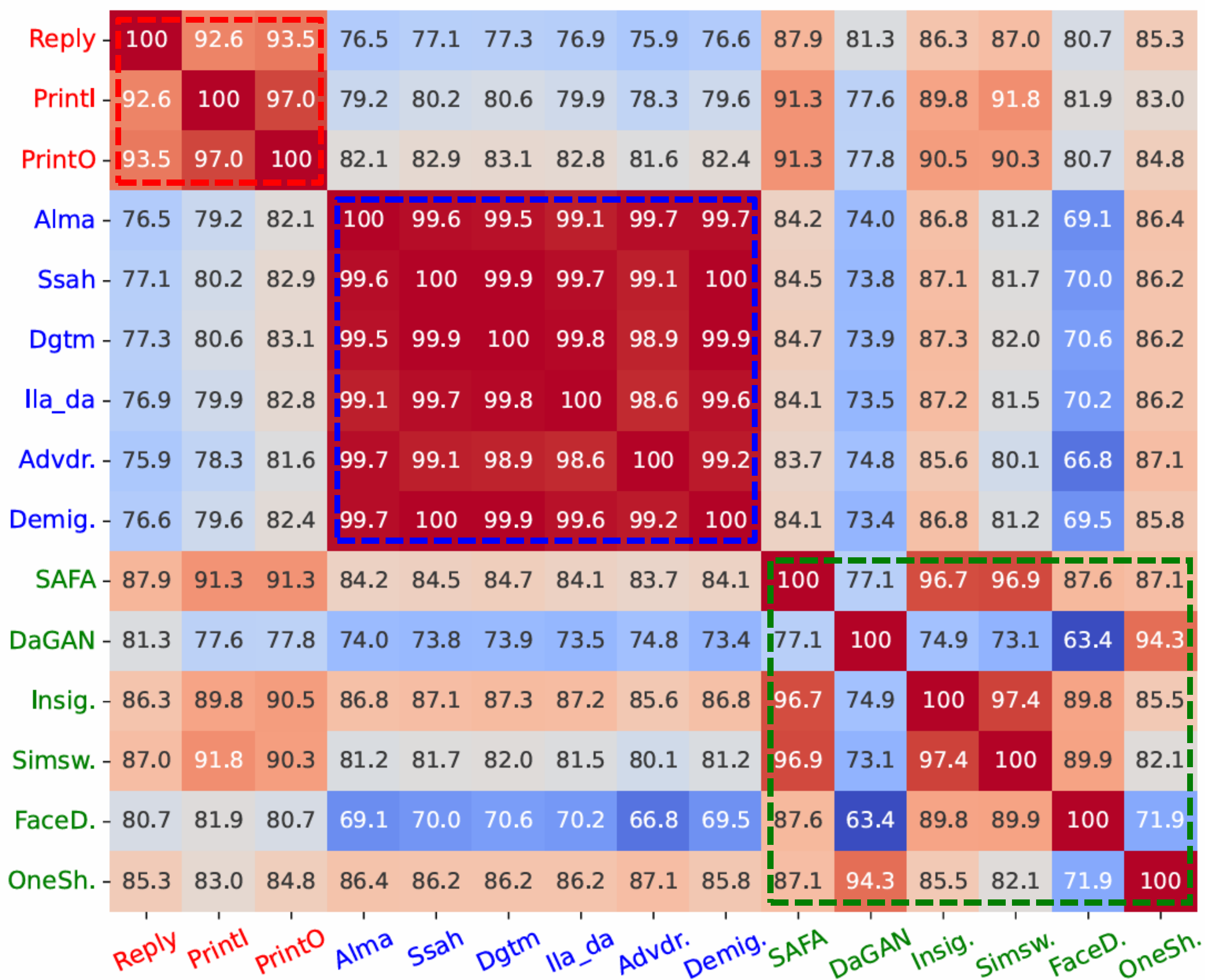}
	\caption{Feature similarity among different attack types (\colorbox{lightred}{Physical}, \colorbox{lightblue}{Advanced}, and \colorbox{lightgreen}{Deepfake}), derived from similarity scores generated by the pre-trained ViT, form the basis for designing more reasonable and challenging protocols.}
	\label{figure3}
\end{figure}

\section{Experiments}
\subsection{Datasets \& Evaluation Metrics}
In this paper, UniAttackData \cite{fang2024unified}  and JFSFDB \cite{yu2024benchmarking} are utilized as experimental datasets. UniAttackData \cite{fang2024unified} is a unified physical-digital attack dataset that covers 3 types of PAs and 12 types of DAs involving 1,800 subjects. It includes 600 subjects / 8400 images for training, 300 subjects / 6000 images for validation, and 900 subjects / 21506 images for testing. JFSFDB \cite{yu2024benchmarking} is the first unified physical and digital face attack detection benchmark that utilizes 9 datasets, including SiW~\cite{liu2018learning}, 3DMAD~\cite{erdogmus2014spoofing}, HKBU-MarsV2~\cite{liu20163d}, MSU-MFSD~\cite{wen2015face}, 3DMask~\cite{yu2020fas}, ROSE-Youtu~\cite{li2018unsupervised}, FaceForensics++~\cite{rossler2019faceforensics++}, DFDC~\cite{dolhansky2019deepfake}, and CelebDFv2~\cite{li2020celeb}. The performance of our model is evaluated with the average classification error rate (ACER), the overall detection accuracy (ACC), the area under curve (AUC), and the equivalent error rate (EER) on UniAttackData. For JFSFDB, the evaluation metrics include EER, AUC, and the single-side True Positive Rate (TPR) at a specified False Positive Rate (FPR). Following the previous works \cite{fang2024unified,zou2024softmoe}, we compare the method with ResNet50, ViT-B/16 \cite{dosovitskiy2020image}, Auxiliary \cite{liu2018learning}, CDCN \cite{yu2020fas}, FFD \cite{dang2020detection}, and UniAttackDetection \cite{fang2024unified}.

\subsection{Implementation Details}
In FA\textsuperscript{3}-CLIP, the image encoder $\mathcal{V}(\cdot)$ is the pre-trained ViT-B/16 \cite{dosovitskiy2020image} and the text encoder is $\mathcal{T}(\cdot)$ the pre-trained Transformer, with $d_{v}=768$, $d_{t}=512$, and $d_{vt}=512$. All parameters are updated during training, except for the text encoder $\mathcal{T}(\cdot)$ which remains frozen. In order to facilitate the learning of a more discriminative textual feature space, a specific training strategy is adopted: only the generic live prompt is generated for live images, while the generic fake prompt is produced in a similar manner for attack images. During the testing phase, both generic live and fake prompts are generated for each image. Futhermore, the number of clusters is set to $d = 32$ in frequency compression block and the hyperparameter $\alpha$ is empirically set to 0.25. The context length is fixed at 6, and its initialization follows the setup used in CoCoOp. Finally, the model is optimized via SGD optimizer with a batch size of 18 and a learning rate of $10^{-6}$.

\begin{table*}
	\caption{Results on the JFSFDB \cite{yu2024benchmarking} with three cross-domain protocols: Face Anti-Spoofing (FAS), Face Forgery Detection (FFD), and Uni-Attack. For FAS, the training set consists of SiW, 3DMAD and HKBU, while the testing set includes MSU, 3DMask and ROSE. For FFD, the training set is FF++, and the testing set comprises DFDC and DFv2. For Uni-Attack, the training set combines SiW, 3DMAD, HKBU and FF++, and the testing set includes MSU, 3DMask, ROSE, DFDC and DFv2. $\downarrow/\uparrow$ indicate that smaller/larger values correspond to better performance. The (Avg.) represents the mean result for all testing sets.}
	\resizebox{\textwidth}{!}{
		\begin{tabular}{|c|c|ccccccccccc|}
			\hline
			&                                   & \multicolumn{11}{c|}{Test Protocols}                                                                                                                                                                                                                                                                                                                                                                                                               \\ \cline{3-13} 
			&                                   & \multicolumn{3}{c}{FAS}                                  & \multicolumn{2}{c|}{FFD}                                  & \multicolumn{5}{c|}{Uni-Attack}                               &              \\ \cline{3-13} 
			\multirow{-3}{*}{Metrics}                    & \multirow{-3}{*}{Methods/Dataset} & \multicolumn{1}{c}{MSU}                                  & \multicolumn{1}{c}{3DMask}                                & \multicolumn{1}{c|}{ROSE}                                     & \multicolumn{1}{c}{DFDC}                                  & \multicolumn{1}{c|}{Celeb-DFv2}                    & \multicolumn{1}{c}{MSU}                                 & \multicolumn{1}{c}{3DMask}                                & \multicolumn{1}{c}{ROSE}                             & \multicolumn{1}{c}{DFDC}                                & \multicolumn{1}{c|}{Celeb-DFv2}                    & Avg.          \\ \hline
			& MesoNet                           & \multicolumn{1}{c}{21.90}                                & \multicolumn{1}{c}{55.82}                                 & \multicolumn{1}{c|}{36.81}                                    & \multicolumn{1}{c}{46.16}                                 & \multicolumn{1}{c|}{38.78}                         & \multicolumn{1}{c}{22.86}                               & \multicolumn{1}{c}{54.76}                                & \multicolumn{1}{c}{36.18}                             & \multicolumn{1}{c}{49.81}                               & \multicolumn{1}{c|}{46.94}                         & 41.00       \\ 
			& Xception                          & \multicolumn{1}{c}{18.57}                                & \multicolumn{1}{c}{42.30}                                 & \multicolumn{1}{c|}{18.13}                                    & \multicolumn{1}{c}{38.92}                                 & \multicolumn{1}{c|}{24.00}                         & \multicolumn{1}{c}{17.14}                               & \multicolumn{1}{c}{\textbf{22.21}}                                & \multicolumn{1}{c}{28.82}                             & \multicolumn{1}{c}{40.75}                               & \multicolumn{1}{c|}{29.16}                         & 28.01       \\ 
			& MultiAtten                        & \multicolumn{1}{c}{18.57}                                & \multicolumn{1}{c}{61.22}                                 & \multicolumn{1}{c|}{33.82}                                    & \multicolumn{1}{c}{39.73}                                 & \multicolumn{1}{c|}{40.46}                         & \multicolumn{1}{c}{13.81}                               & \multicolumn{1}{c}{44.42}                                & \multicolumn{1}{c}{30.32}                             & \multicolumn{1}{c}{43.93}                               & \multicolumn{1}{c|}{43.57}                         & 36.99       \\ 
			& CDCN++                            & \multicolumn{1}{c}{22.80}                                & \multicolumn{1}{c}{52.76}                                 & \multicolumn{1}{c|}{32.01}                                    & \multicolumn{1}{c}{44.75}                                 & \multicolumn{1}{c|}{28.19}                         & \multicolumn{1}{c}{31.90}                               & \multicolumn{1}{c}{43.83}                                & \multicolumn{1}{c}{33.19}                             & \multicolumn{1}{c}{44.58}                               & \multicolumn{1}{c|}{29.68}                         & 36.37       \\ 
			& DeepPixel                         & \multicolumn{1}{c}{15.71}                                & \multicolumn{1}{c}{46.42}                                 & \multicolumn{1}{c|}{28.23}                                    & \multicolumn{1}{c}{36.82}                                 & \multicolumn{1}{c|}{\textbf{22.81}}                & \multicolumn{1}{c}{13.33}                               & \multicolumn{1}{c}{40.19}                                & \multicolumn{1}{c}{32.80}                             & \multicolumn{1}{c}{37.38}                               & \multicolumn{1}{c|}{19.48}                         & 29.32       \\ 
			& ResNet50                          & \multicolumn{1}{c}{12.38}                                & \multicolumn{1}{c}{45.24}                                 & \multicolumn{1}{c|}{19.94}                                    & \multicolumn{1}{c}{40.61}                                 & \multicolumn{1}{c|}{26.01}                         & \multicolumn{1}{c}{12.86}                               & \multicolumn{1}{c}{28.79}                                & \multicolumn{1}{c}{21.55}                             & \multicolumn{1}{c}{36.82}                               & \multicolumn{1}{c|}{23.06}                         & 26.73       \\ 
			& ViT                               & \multicolumn{1}{c}{11.90}                                & \multicolumn{1}{c}{60.40}                                 & \multicolumn{1}{c|}{19.47}                                    & \multicolumn{1}{c}{32.75}                                 & \multicolumn{1}{c|}{27.76}                         & \multicolumn{1}{c}{9.52}                                & \multicolumn{1}{c}{46.30}                                & \multicolumn{1}{c}{16.20}                             & \multicolumn{1}{c}{30.99}                               & \multicolumn{1}{c|}{20.01}                         & 27.53        \\ 
			& VIT(shared 8)                     & \multicolumn{1}{c}{-}                                    & \multicolumn{1}{c}{-}                                     & \multicolumn{1}{c|}{-}                                        & \multicolumn{1}{c}{-}                                     & \multicolumn{1}{c|}{-}                             & \multicolumn{1}{c}{6.67}                                 & \multicolumn{1}{c}{43.60}                                & \multicolumn{1}{c}{11.64}                             & \multicolumn{1}{c}{30.60}                               & \multicolumn{1}{c|}{\textbf{18.80}}                & 22.23       \\ 
			\multirow{-9}{*}{EER(\%) $\downarrow$}       & \cellcolor{gray!30}Ours           & \multicolumn{1}{c}{\cellcolor{gray!30}\textbf{6.90}}     & \multicolumn{1}{c}{\cellcolor{gray!30}\textbf{18.87}}     & \multicolumn{1}{c|}{\cellcolor{gray!30}\textbf{9.61}}         & \multicolumn{1}{c}{\cellcolor{gray!30}\textbf{22.46}}     & \multicolumn{1}{c|}{\cellcolor{gray!30}25.51}      & \multicolumn{1}{c}{\cellcolor{gray!30}{\textbf{4.29}}}  & \multicolumn{1}{c}{\cellcolor{gray!30}24.00}             & \multicolumn{1}{c}{\cellcolor{gray!30}\textbf{10.77}} & \multicolumn{1}{c}{\cellcolor{gray!30}\textbf{24.53}}   & \multicolumn{1}{c|}{\cellcolor{gray!30}30.02}      & \cellcolor{gray!30}\textbf{17.69}       \\ \hline
			& MesoNet                           & \multicolumn{1}{c}{85.33}                                & \multicolumn{1}{c}{44.54}                                 & \multicolumn{1}{c|}{68.05}                                    & \multicolumn{1}{c}{54.72}                                 & \multicolumn{1}{c|}{65.10}                         & \multicolumn{1}{c}{84.01}                               & \multicolumn{1}{c}{45.87}                                & \multicolumn{1}{c}{70.00}                             & \multicolumn{1}{c}{50.57}                               & \multicolumn{1}{c|}{55.03}                         & 62.32        \\ 
			& Xception                          & \multicolumn{1}{c}{90.62}                                & \multicolumn{1}{c}{63.88}                                 & \multicolumn{1}{c|}{90.15}                                    & \multicolumn{1}{c}{64.86}                                 & \multicolumn{1}{c|}{84.27}                         & \multicolumn{1}{c}{91.56}                               & \multicolumn{1}{c}{75.49}                                & \multicolumn{1}{c}{79.38}                             & \multicolumn{1}{c}{63.69}                               & \multicolumn{1}{c|}{78.87}                         & 78.28       \\ 
			& MultiAtten                        & \multicolumn{1}{c}{89.21}                                & \multicolumn{1}{c}{36.88}                                 & \multicolumn{1}{c|}{72.66}                                    & \multicolumn{1}{c}{63.37}                                 & \multicolumn{1}{c|}{64.34}                         & \multicolumn{1}{c}{92.89}                               & \multicolumn{1}{c}{58.31}                                & \multicolumn{1}{c}{77.51}                             & \multicolumn{1}{c}{58.11}                               & \multicolumn{1}{c|}{60.00}                         & 67.33       \\ 
			& CDCN++                            & \multicolumn{1}{c}{82.65}                                & \multicolumn{1}{c}{47.66}                                 & \multicolumn{1}{c|}{76.76}                                    & \multicolumn{1}{c}{56.65}                                 & \multicolumn{1}{c|}{78.34}                         & \multicolumn{1}{c}{78.93}                               & \multicolumn{1}{c}{61.54}                                & \multicolumn{1}{c}{76.39}                             & \multicolumn{1}{c}{56.42}                               & \multicolumn{1}{c|}{76.92}                         & 69.23       \\ 
			& DeepPixel                         & \multicolumn{1}{c}{95.17}                                & \multicolumn{1}{c}{56.87}                                 & \multicolumn{1}{c|}{80.6}                                     & \multicolumn{1}{c}{67.54}                                 & \multicolumn{1}{c|}{85.51}                         & \multicolumn{1}{c}{93.97}                               & \multicolumn{1}{c}{66.65}                                & \multicolumn{1}{c}{74.58}                             & \multicolumn{1}{c}{66.36}                               & \multicolumn{1}{c|}{88.45}                         & 77.57        \\
			& ResNet50                          & \multicolumn{1}{c}{94.33}                                & \multicolumn{1}{c}{58.95}                                 & \multicolumn{1}{c|}{88.79}                                    & \multicolumn{1}{c}{62.35}                                 & \multicolumn{1}{c|}{82.51}                         & \multicolumn{1}{c}{93.50}                               & \multicolumn{1}{c}{79.64}                                & \multicolumn{1}{c}{86.03}                             & \multicolumn{1}{c}{67.18}                               & \multicolumn{1}{c|}{85.55}                         & 79.88       \\ 
			& ViT                               & \multicolumn{1}{c}{95.13}                                & \multicolumn{1}{c}{35.60}                                 & \multicolumn{1}{c|}{89.12}                                    & \multicolumn{1}{c}{73.79}                                 & \multicolumn{1}{c|}{\textbf{86.61}}                & \multicolumn{1}{c}{95.85}                               & \multicolumn{1}{c}{62.74}                                & \multicolumn{1}{c}{91.88}                             & \multicolumn{1}{c}{75.11}                               & \multicolumn{1}{c|}{88.36}                         & 79.42       \\ 
			& VIT(shared 8)                     & \multicolumn{1}{c}{-}                                    & \multicolumn{1}{c}{-}                                     & \multicolumn{1}{c|}{-}                                        & \multicolumn{1}{c}{-}                                     & \multicolumn{1}{c|}{-}                             & \multicolumn{1}{c}{97.99}                               & \multicolumn{1}{c}{67.10}                                & \multicolumn{1}{c}{95.57}                             & \multicolumn{1}{c}{75.90}                               & \multicolumn{1}{c|}{\textbf{89.76}}                & 85.26       \\ 
			\multirow{-9}{*}{AUC(\%) $\uparrow$}         & \cellcolor{gray!30}Ours           & \multicolumn{1}{c}{\cellcolor{gray!30}\textbf{97.96}}    & \multicolumn{1}{c}{\cellcolor{gray!30}\textbf{89.99}}     & \multicolumn{1}{c|}{\cellcolor{gray!30}\textbf{95.52}}        & \multicolumn{1}{c}{\cellcolor{gray!30}\textbf{85.60}}     & \multicolumn{1}{c|}{\cellcolor{gray!30}83.15}      & \multicolumn{1}{c}{\cellcolor{gray!30}\textbf{98.12}}   & \multicolumn{1}{c}{\cellcolor{gray!30}\textbf{81.84}}    & \multicolumn{1}{c}{\cellcolor{gray!30}\textbf{95.58}} & \multicolumn{1}{c}{\cellcolor{gray!30}\textbf{82.95}}   & \multicolumn{1}{c|}{\cellcolor{gray!30}74.80}                         & \cellcolor{gray!30}\textbf{88.55}       \\ \hline
			& MesoNet                           & \multicolumn{3}{c|}{28.72, 9.10}                         & \multicolumn{2}{c|}{24.92, 2.71}                          & \multicolumn{5}{c|}{24.85, 5.77}                              & 25.83, 5.86  \\ 
			& Xception                          & \multicolumn{3}{c|}{55.97, 30.83}                        & \multicolumn{2}{c|}{53.46, 11.19}                         & \multicolumn{5}{c|}{44.58, 16.99}                             & 51.33, \textbf{59.01} \\ 
			& MultiAtten                        & \multicolumn{3}{c|}{34.13, 13.46}                        & \multicolumn{2}{c|}{32.15, 3.90}                          & \multicolumn{5}{c|}{36.88, 11.37}                             & 34.39, 9.58  \\ 
			& CDCN++                            & \multicolumn{3}{c|}{36.51, 18.04}                        & \multicolumn{2}{c|}{51.34, 12.46}                         & \multicolumn{5}{c|}{45.43, 17.45}                             & 44.42, 15.98 \\ 
			& DeepPixel                         & \multicolumn{3}{c|}{40.62, 22.09}                        & \multicolumn{2}{c|}{52.42, 9.34}                          & \multicolumn{5}{c|}{49.98, 43.52}                             & 47.67, 24.98 \\ 
			& ResNet50                          & \multicolumn{3}{c|}{53.61, 29.16}                        & \multicolumn{2}{c|}{50.43, 8.51}                          & \multicolumn{5}{c|}{57.66, \textbf{48.35}}                    & 53.90, 28.67  \\ 
			& ViT                               & \multicolumn{3}{c|}{52.69, 29.80}                        & \multicolumn{2}{c|}{56.41, 10.31}                         & \multicolumn{5}{c|}{61.67, 22.60}                             & 56.92, 20.90 \\ 
			& VIT(shared 8)                     & \multicolumn{3}{c|}{-}                                   & \multicolumn{2}{c|}{-}                                    & \multicolumn{5}{c|}{\textbf{68.32}, 26.52}                    & 68.32, 26.52 \\ 
			\multirow{-9}{*}{\parbox{3.3cm}{TPR(\%)@FPR=10\% $\uparrow$,  
					\\ TPR(\%)@FPR=1\% $\uparrow$}}              & \cellcolor{gray!30}Ours   & \multicolumn{3}{c|}{\cellcolor{gray!30}\textbf{86.67, 34.7}}     & \multicolumn{2}{c|}{\cellcolor{gray!30}\textbf{58.11, 15.74}}        & \multicolumn{5}{c|}{\cellcolor{gray!30}64.68, 16.34}     & \cellcolor{gray!30}\textbf{69.82}, 22.26 \\ \hline
		\end{tabular}
	}
	\label{tab2}
\end{table*}

\begin{table}[h]
	\caption{Results on the UniAttackData \cite{fang2024unified} with standard \emph{Protocol 1} and our  \emph{Protocols 1.1, 1.2} and \emph{1.3}. $\downarrow/\uparrow$ indicate that smaller/larger values correspond to better performance. The (Avg.) represents the mean result for all protocols.}
	\centering
	\resizebox{\linewidth}{!}{
		\begin{tabular}{c|c|cccc|c}
			\hline
			& \multicolumn{1}{c|}{}                          & \multicolumn{5}{c}{Protocols}                                                                                                                                                                            \\ \cline{3-7} 
			\multirow{-2}{*}{Metrics} & \multicolumn{1}{c|}{\multirow{-2}{*}{Methods}}                   & \multicolumn{1}{c}{P1}                 & \multicolumn{1}{c}{P1.1}              & \multicolumn{1}{c}{P1.2}               & \multicolumn{1}{c}{P1.3}               & \multicolumn{1}{|c}{Avg.}      \\ \hline
			& CLIP~\cite{radford2021learning}                                      & 1.02                                   & 14.81                                 & 5.36                                   & 2.45                                   & 5.91                                   \\
			& CDCN~\cite{yu2020searching}                                      & 1.40                                   & 12.32                                 & 16.34                                  & 4.41                                   & 8.62                                   \\
			& VIT-B/16~\cite{dosovitskiy2020image}                             & 5.92                                   & 13.53                                 & 5.22                                   & 3.20                                   & 6.97                                   \\
			& ResNet50~\cite{he2016deep}                                       & 1.35                                   & \textbf{5.92}                         & 25.90                                  & 4.92                                   & 9.52                                   \\
			& UniAttackDetection~\cite{fang2024unified}                        & 0.52                                   & 11.73                                 & 1.70                                   & 4.67                                   & 4.66                                   \\
			\multirow{-6}{*}{ACER(\%)$\downarrow$}    & \cellcolor{gray!30}\textbf{Ours}                 & \cellcolor{gray!30}\textbf{0.36}       & \cellcolor{gray!30}9.57               & \cellcolor{gray!30}\textbf{1.43}       & \cellcolor{gray!30}\textbf{2.30}        & \cellcolor{gray!30}\textbf{3.42}  \\ \cline{1-7}
			& CLIP~\cite{radford2021learning}                                      & 99.01                                  & 74.27                                 & 92.48                                  & 96.93                                  & 90.67                                  \\
			& CDCN~\cite{yu2020searching}                                      & 98.57                                  & 86.28                                 & 79.36                                  & 94.90                                  & 89.78                                  \\
			& VIT-B/16~\cite{dosovitskiy2020image}                             & 92.29                                  & 80.98                                 & 92.72                                  & 95.72                                  & 90.43                                  \\
			& ResNet50~\cite{he2016deep}                                       & 98.83                                  & 76.64                                 & 64.35                                  & 94.05                                  & 83.47                                  \\
			& UniAttackDetection~\cite{fang2024unified}                        & 99.45                                  & 83.56                                 & 97.88                                  & 93.85                                  & 93.69                                  \\
			\multirow{-6}{*}{ACC(\%)$\uparrow$ }     & \cellcolor{gray!30}\textbf{Ours}                  & \cellcolor{gray!30}\textbf{99.56}      & \cellcolor{gray!30}\textbf{86.6}      & \cellcolor{gray!30}\textbf{98.27}      & \cellcolor{gray!30}\textbf{97.25}      & \cellcolor{gray!30}\textbf{95.42} \\ \cline{1-7}
			& CLIP~\cite{radford2021learning}                                      & 99.47                                  & 86.74                                 & 99.17                                  & 97.92                                  & 95.83                                  \\
			& CDCN~\cite{yu2020searching}                                      & 99.52                                  & 93.89                                 & 93.34                                  & 97.68                                  & 96.11                                  \\
			& VIT-B/16~\cite{dosovitskiy2020image}                             & 97.00                                  & 95.99                                 & 99.36                                  & 99.18                                  & 97.88                                  \\
			& ResNet50~\cite{he2016deep}                                       & 99.79                                  & 91.25                                 & 84.35                                  & 98.84                                  & 93.56                                  \\
			& UniAttackDetection~\cite{fang2024unified}                        & \textbf{99.95}                         & \textbf{98.81}                        & 99.85                                  & 99.13                                  & \textbf{99.44}                         \\
			\multirow{-6}{*}{AUC(\%)$\uparrow$}     & \cellcolor{gray!30}\textbf{Ours}                   & \cellcolor{gray!30}99.75               & \cellcolor{gray!30}96.49              & \cellcolor{gray!30}\textbf{99.85}      & \cellcolor{gray!30}\textbf{99.19}      & \cellcolor{gray!30}98.82          \\ \cline{1-7}
			& CLIP~\cite{radford2021learning}                                      & 0.96                                   & 26.45                                 & 2.06                                   & 4.11                                   & 10.87                                  \\
			& CDCN~\cite{yu2020searching}                                      & 1.42                                   & 12.75                                 & 13.84                                  & 6.28                                   & 8.57                                   \\
			& VIT-B/16~\cite{dosovitskiy2020image}                             & 9.14                                   & 10.98                                 & 2.23                                   & 2.92                                   & 6.32                                   \\
			& ResNet50~\cite{he2016deep}                                       & 1.18                                   & 16.70                                 & 23.26                                  & 5.06                                   & 11.55                                  \\
			& UniAttackDetection~\cite{fang2024unified}                        & 0.53                                   & 7.40                                  & 1.75                                   & 4.25                                   & 3.48                                   \\
			\multirow{-6}{*}{EER(\%)$\downarrow$}     & \cellcolor{gray!30}\textbf{Ours}                 & \cellcolor{gray!30}\textbf{0.49}       & \cellcolor{gray!30}\textbf{6.52}      & \cellcolor{gray!30}\textbf{1.60}       & \cellcolor{gray!30}\textbf{2.26}       & \cellcolor{gray!30}\textbf{2.72}  \\ \cline{1-7}
		\end{tabular}
	}
	\label{tab3}
\end{table}

\subsection{Experimental Result}
\noindent {\bf{Comparisons on UniAttackData.}} Four protocols are used to evaluate our method on UniAttackData \cite{fang2024unified}, with results summarized in Table~\ref{tab3}. Under \emph{Protocol 1}, our method achieves the best performance on most metrics, confirming its effectiveness. However, results approaching theoretical limits accompanied by overfitting and identity leakage indicate that the original protocol is inadequate for accurately assessing model performance. To address these concerns, we introduce new protocols characterized by stricter identity partitioning and more challenging data splits. As anticipated, these modifications result in performance declines across all metrics, highlighting increased evaluation difficulty. Nevertheless, our method consistently achieves state-of-the-art (SOTA) performance under \emph{Protocol 1.2} and \emph{Protocol 1.3}, further validating its robustness and effectiveness.

Notably, performance (ACER and AUC) under \emph{Protocol 1.1} is lower compared to other protocols, primarily attributable to the distinctive characteristics of the deepfake subtype (illustrated in Fig.~\ref{figure3}). The pronounced divergence in latent representations indicates that deepfake data constitutes a more challenging detection benchmark, indirectly validating the capability of FA\textsuperscript{3}-CLIP in capturing subtle fake cues.

\noindent {\bf{Comparisons on JFSFDB.}} We further evaluate our method on the JFSFDB dataset \cite{yu2024benchmarking}, with results compared against other models in Table~\ref{tab2}. Although our method achieves state-of-the-art performance across most metrics, the lower scores observed under the Uni-Attack protocol at TPR@FPR=1\% highlight the challenges posed by stricter evaluation criteria, where precision becomes paramount. The Uni-Attack protocol, encompassing a diverse array of attack types, represents a more complex scenario demanding higher precision to sustain low false-positive rates, explaining the performance decline at the stricter FPR=1\% threshold. Despite this, overall results demonstrate the robustness and effectiveness of our approach, with the relatively lower TPR@FPR=1\% highlighting areas for potential improvement.

\begin{table}[!h]
	\caption{Ablation Results on Dual-Stream Cues Fusion Framework (Dual.), Attack-Agnostic Prompt (Prompt.) and $\mathcal{L}_{nt}$ with \emph{Protocol 1} on UniAttackData \cite{fang2024unified}. $\downarrow/\uparrow$ indicate that smaller/larger values correspond to better performance.}
	\centering
	\resizebox{\linewidth}{!}{
		\scalebox{1.0}{
			\begin{tabular}{c c c c | c c c c}
				\hline
				CLIP         & Dual.            & Prompt.           & $\mathcal L_{nt}$  &ACER(\%)$\downarrow$ &ACC(\%)$\uparrow$ &AUC(\%)$\uparrow$ &EER(\%)$\downarrow$\\
				\hline
				\usym{1F5F8} & -                 & -                  & -                    & 1.02          & 99.01          & 99.47   & 0.96\\
				\usym{1F5F8} & \usym{1F5F8}      & -                  & -                    & 0.59          & 99.41          & 99.70   & 0.65\\
				\usym{1F5F8} & -                 & \usym{1F5F8}       & -                    & 0.66          & 97.72          & 99.62   & 0.71\\
				\usym{1F5F8} & -                 & \usym{1F5F8}       & \usym{1F5F8}         & 0.44          & 99.49          & 99.72   & 0.57\\
				\rowcolor{gray!30}
				\usym{1F5F8} & \usym{1F5F8}      & \usym{1F5F8}       & \usym{1F5F8}    & \textbf{0.36} & \textbf{99.56} & \textbf{99.75}  & \textbf{0.49}\\
				\hline
	\end{tabular}}}
	\label{tab4}
\end{table}

\subsection{Ablation Experiments}
\begin{figure}[!h]
	\centering
	\includegraphics[width=1.0\linewidth]{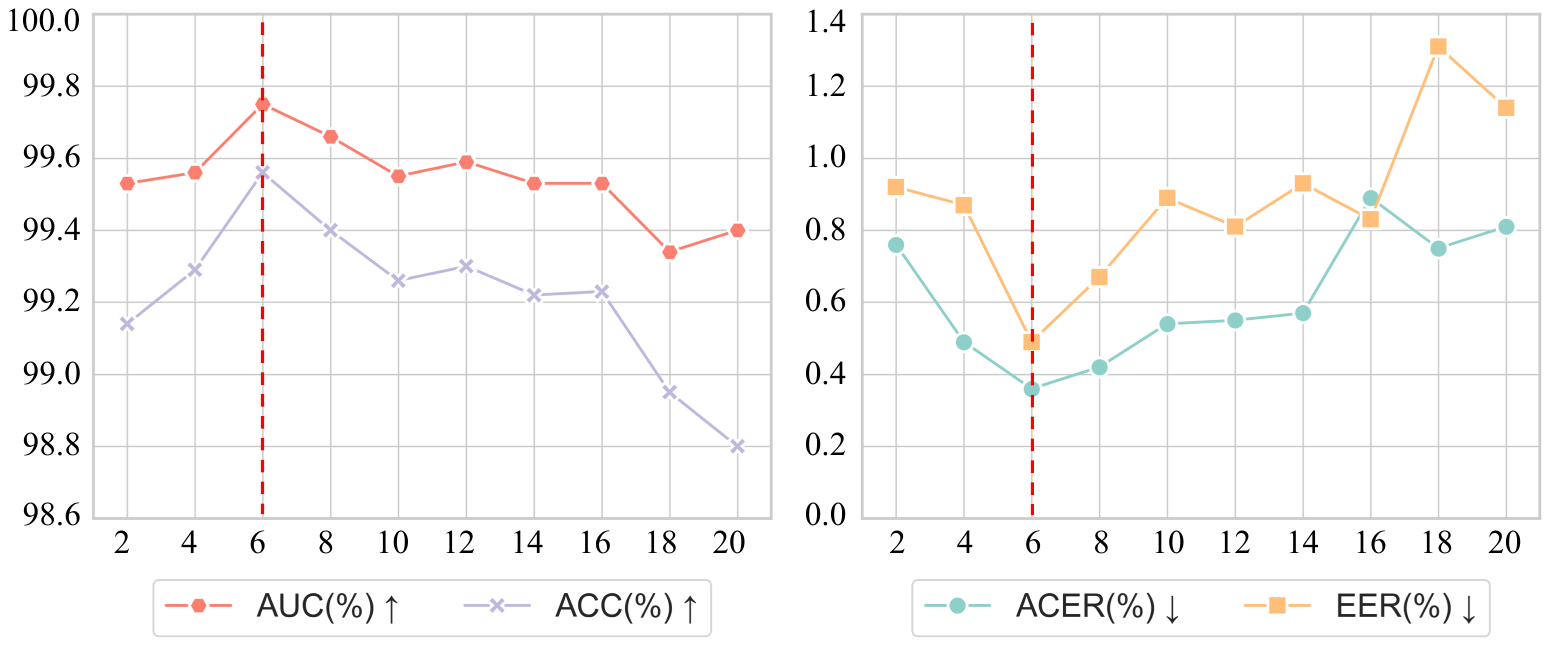}
	\caption{Ablation results on UniAttackData \cite{fang2024unified} with \emph{Protocol 1}. $\downarrow/\uparrow$ indicate that smaller/larger values correspond to better performance. The value of the context length performs well in 6.}
	\label{figure4}
\end{figure}

We conduct ablation studies on both the language and vision branches using the UniAttackData \cite{fang2024unified} with \emph{Protocol 1}. The outcomes of these experiments are detailed in Fig.~\ref{figure4}, Table~\ref{tab4}, Table~\ref{tab5}, and Table~\ref{tab6}.

\noindent {\bf{Ablation on each branch with FA\textsuperscript{3}-CLIP.}} To evaluate the contributions of each proposed component in FA\textsuperscript{3}-CLIP, such as the dual-stream cues fusion framework, attack-agnostic prompt, and $\mathcal{L}_{nt}$, we incrementally integrate them into the baseline CLIP \cite{radford2021learning}, with results summarized in Table~\ref{tab4}. Introducing the dual-stream fusion framework significantly improves model performance, demonstrating the effectiveness of incorporating frequency information as complementary cues, thereby validating our hypothesis that frequency features substantially differ between live and fake faces (illustrated in Fig.~\ref{figure1}). 

\begin{table}[h]
	\caption{
		Ablation Results on Attack-Agnostic Prompt with Original Spatial and frequency Input Features on UniAttackData \cite{fang2024unified} using \emph{Protocol 1}. $\downarrow/\uparrow$ indicate that smaller/larger values correspond to better performance.
	}
	\centering
	\resizebox{\linewidth}{!}{
		\begin{tabular}{c c c|c c c c}
			\hline
			Context     & Spatial           & Freq          &ACER(\%)$\downarrow$ &ACC(\%)$\uparrow$ &AUC(\%)$\uparrow$ &EER(\%)$\downarrow$ \\
			\hline
			\usym{1F5F8} & -             & -             & 0.64          & 99.27          & 99.60          & 0.79 \\
			\usym{1F5F8} & \usym{1F5F8}  & -             & 0.67          & 99.24          & 99.58          & 0.83 \\
			\usym{1F5F8} & -             & \usym{1F5F8}  & 0.54          & 99.39          & 99.66          & 0.66 \\
			
			\rowcolor{gray!30}
			\usym{1F5F8} & \usym{1F5F8}  & \usym{1F5F8}  &\textbf{0.36}  &\textbf{99.56}  &\textbf{99.75}  &\textbf{0.49} \\
			
			\hline
			
	\end{tabular}}
	\label{tab5}
\end{table}

We also observe that generic prompts offer fewer benefits than frequency information, which can be attributed to the fact that relying solely on textual “live” and “fake” labels provides limited discriminative cues. Further introducing the normalized temperature-scaled cross-entropy loss $\mathcal{L}_{nt}$ optimizes these generic live and fake prompts by enforcing intra-class compactness and inter-class separation, significantly enhancing model performance. Integrating all proposed enhancements into the baseline yields the best results, demonstrating stability and positive synergy among the introduced components.

\noindent {\bf{Ablation on the attack-agnostic prompt.}} We perform detailed ablation studies on the language branch, examining the influence of various input types on attack-agnostic prompt generation, as presented in Table~\ref{tab5}. When relying solely on spatial features from images for prompt generation, the performance is marginally inferior compared to using learnable context. This gap may result from spatial features being insufficient to clearly differentiate live and fake cues within textual feature space. In contrast, leveraging frequency information for generic prompt generation results in a significant performance improvement, underscoring that frequency components in facial images capture more discriminative live and fake cues. Moreover, the integration of both spatial and frequency features to generate generic prompt achieves optimal performance, highlighting that spatial cues alone are insufficient and must be complemented by frequency features to maximize the model's performance. Additionally, ablation studies regarding context length (as shown in Fig.~\ref{figure4}) indicate that a length of 6 achieves optimal performance by balancing expressiveness and redundancy.

\begin{table}[!h]
	\caption{Ablation Results on frequency Features using UniAttackData \cite{fang2024unified} and \emph{Protocol 1}. The Impact of the Frequency Compression Block (FCB), Multi-layer frequency, and Original Image Frequency. $\downarrow/\uparrow$ indicate that smaller/larger values correspond to better performance.}
	\centering
	\resizebox{\linewidth}{!}{
		\scalebox{1.0}{
			\begin{tabular}{c c c | c c c c}
				\hline
				FCB          & MultiFreq          & OriFreq            &ACER(\%)$\downarrow$  &ACC(\%)$\uparrow$ &AUC(\%)$\uparrow$ &EER(\%)$\downarrow$ \\
				\hline
				-            & \usym{1F5F8}      & \usym{1F5F8}       & 0.67                 & 99.26            & 99.59            & 0.79 \\
				\usym{1F5F8} & -                 & \usym{1F5F8}       & 0.48                 & 99.30            & 99.57            & 0.86 \\
				\usym{1F5F8} & \usym{1F5F8}      & -                  & 0.45                 & 99.47            & 99.70            & 0.59 \\
				\rowcolor{gray!30}
				\usym{1F5F8} & \usym{1F5F8}      & \usym{1F5F8}       & \textbf{0.36}        & \textbf{99.56}   & \textbf{99.75}   & \textbf{0.49}\\
				\hline
	\end{tabular}}}
	\label{tab6}
\end{table}


\begin{figure*}[!t]
	\centering
	\includegraphics[width=1.0\linewidth]{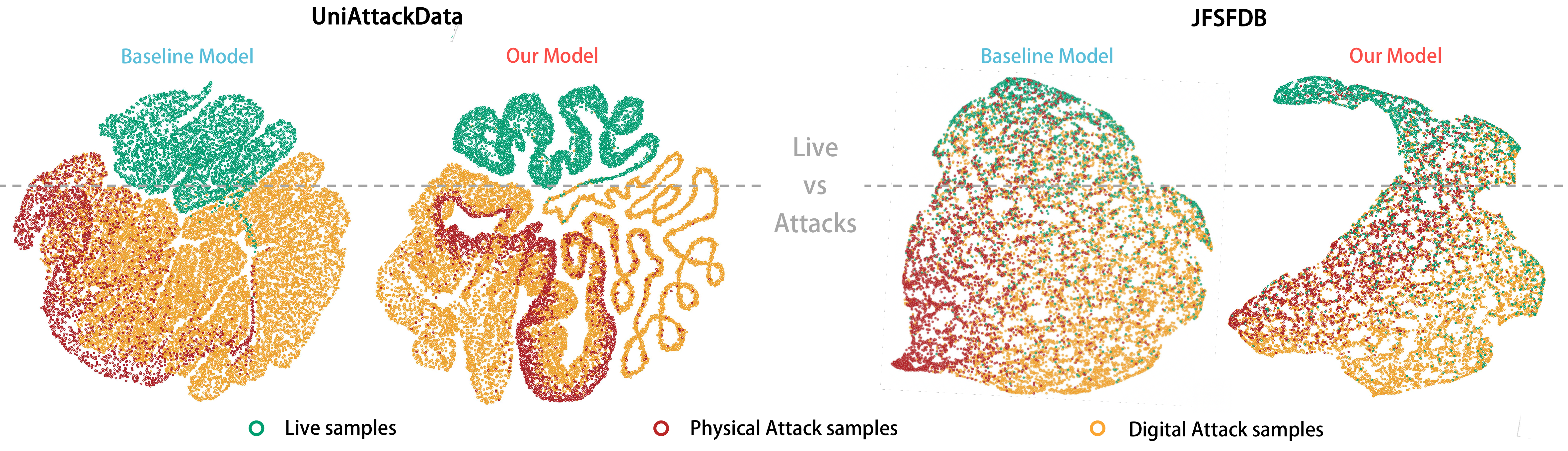}
	\caption{UMAP \cite{mcinnes2018umap} visualization of the feature representations learned by baseline CLIP and our model on both UniAttackData \cite{fang2024unified} and JFSFDB \cite{yu2024benchmarking}. Compared to CLIP, our method yields more distinctly separated clusters for live faces and fake faces.}
	\label{figure5}
\end{figure*}

\begin{figure}[!h]
	\centering
	\includegraphics[width=1.0\linewidth]{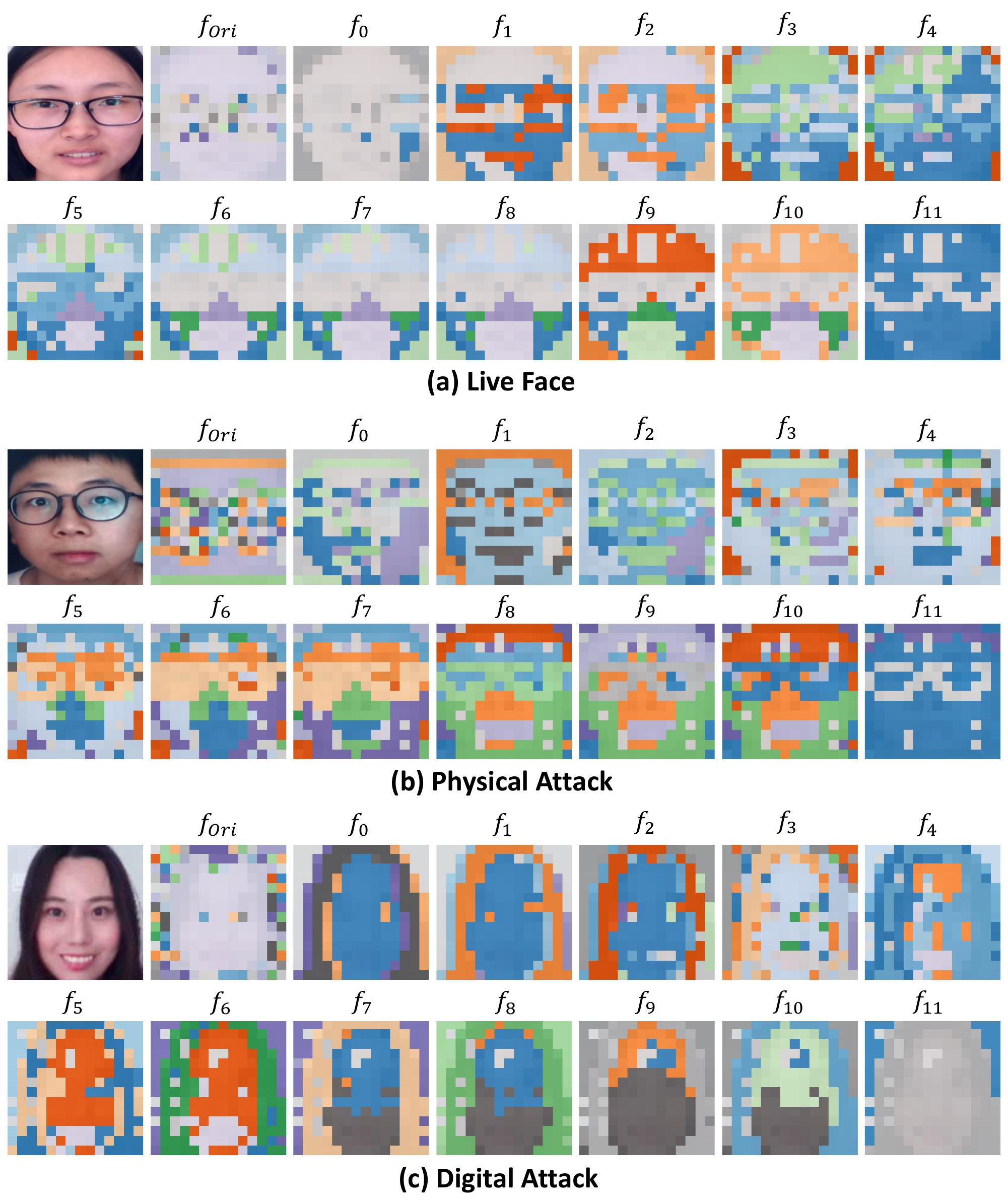}
	\caption{The visualization reveals that different layers exhibit diverse critical cues and substantial redundancy in frequency-domain features, confirming the theoretical findings presented in Section \ref{DSCFF_subsection}. $\boldsymbol{f}_{Ori}$ denotes the frequency-aware features of original image, the others are the multi-layer frequency-aware features from ViT.}
	\label{figure6}
\end{figure}

\noindent {\bf{Ablation on the frequency features.}} To validate the effectiveness of frequency information, we perform ablation experiments on the vision branch. As illustrated in Table~\ref{tab6}, integrating frequency features from both original image and multi-layer features into the spatial domain results in a performance degradation. This decline can be attributed to an excess of redundant frequency patches, which obscure critical cues from a smaller subset of valuable patches. 

However, after incorporating the frequency compression block (FCB) to compress these frequency cues, the model's performance improves significantly. This result indicates that redundant information is indeed present among the multi-layer frequency features (supported by Fig.~\ref{figure6}). Additionally, we observe that utilizing multi-layer frequency features outperforms using the frequency information from the original image, achieving a 0.27\% improvement in EER, suggesting richer discriminative cues provided by multi-layer frequency features. Ultimately, the results validate that frequency cues effectively complement spatial information.

\subsection{Visualization Analysis}
Our algorithm leverages correlations among multi-layer frequency cues from the vision branch. To investigate model behavior, we visualize frequency cues identified at each Transformer layer. Representative images from each category processed by FA\textsuperscript{3}-CLIP are shown in Fig.~\ref{figure6}, where patches matched to similar cues are highlighted in the same color. We observe abstract features captured in shallow network layers, while deeper layers progressively extract concrete frequency cues corresponding to visually detectable attack regions, highlighting the model's capability to effectively identify and distinguish live and fake cues.

To assess the separability of live and fake face embeddings, we utilize UMAP \cite{mcinnes2018umap} for visualization (Fig.~\ref{figure5}). Embeddings for live faces, physical attacks, and digital attacks are represented by green, red, and yellow points, respectively. Our method achieves clear and robust separation, with live face embeddings tightly clustered within a distinct region.

\section{Conclusion}
We propose FA\textsuperscript{3}-CLIP to effectively detect both physical and digital facial attacks within a unified feature space. Specifically, we introduce attack-agnostic prompt learning to generate generic live and fake prompts, enabling effective discrimination of live face and diverse facial attacks. Furthermore, we design a dual-stream cue fusion framework that incorporates frequency information extracted from multiple transformer layers to complement spatial features. A number of quantitative and visual experiments demonstrate the effectiveness of our theories. Additionally, we propose rigorous protocols to to facilitate unified face attack detection effectiveness. In future work, we will focus on improving the computational efficiency of the dual-stream feature fusion process. Moreover, developing more adaptive frequency filtering mechanisms to refine the extraction of discriminative frequency cues.

\bibliographystyle{IEEEtran}
\bibliography{IEEEabrv, main}



\end{document}